\documentclass[arxiv]{article}

\usepackage[utf8]{inputenc} 
\usepackage[T1]{fontenc}    
\usepackage{hyperref}       
\usepackage{url}            
\usepackage{booktabs}       
\usepackage{amsfonts}       
\usepackage{nicefrac}       
\usepackage{microtype}      
\usepackage{xcolor}         
\usepackage{wrapfig}
\usepackage{caption}
\usepackage{soul}
\usepackage{url}
\usepackage{epstopdf}
\usepackage{float}
\usepackage[utf8]{inputenc}
\usepackage{graphicx}
\usepackage{amsmath}
\usepackage{amssymb}
\usepackage{booktabs}
\usepackage{algorithm}
\usepackage[noend]{algorithmic}
\usepackage{enumitem}
\usepackage{booktabs} 
\usepackage{graphicx,array}
\usepackage{color,soul,xspace}

\usepackage{comment}
\usepackage{epsfig}
\usepackage{subcaption}
\usepackage{mathrsfs,mdwlist,enumitem}

\newcommand{\name}{\textsc{LGnn}\xspace}
\newcommand{\names}{\textsc{LGnn}s\xspace}

\newcommand{\lnn}{\textsc{Lnn}\xspace}
\newcommand{\hnn}{\textsc{Hnn}\xspace}
\newcommand{\hnns}{\textsc{Hnn}s\xspace}
\newcommand{\lnns}{\textsc{Lnn}s\xspace}
\newcommand{\MLP}{\texttt{MLP}\xspace}
\newcommand{\sqp}{\texttt{squareplus}\xspace}

\newcommand{\CU}{\mathcal{U}\xspace}

\newcommand{\CE}{\mathcal{E}\xspace}

\newcommand{\cW}{\mathbf{W}\xspace}

\newcommand{\ch}{\mathbf{h}\xspace}

\newcommand{\cz}{\mathbf{z}\xspace}

\setlist{nolistsep,leftmargin=*}

\newtheorem{thm}{\textbf{Theorem}}

\title{Learning the Dynamics of Particle-based Systems with Lagrangian Graph Neural Networks}

\author{
Ravinder Bhattoo\thanks{
Department of Civil Engineering, Indian Institute of Technology Delhi, Hauz Khas, New Delhi, India 110016, \texttt{ravinder@iitd.ac.in}},
Sayan Ranu\thanks{Department of Computer Science and Engineering, Indian Institute of Technology Delhi, Hauz Khas, New Delhi, India 110016, \texttt{sayanranu@iitd.ac.in}},
N. M. Anoop Krishnan\thanks{
Department of Civil Engineering, Indian Institute of Technology Delhi, Hauz Khas, New Delhi, India 110016, \texttt{krishnan@iitd.ac.in}},
}

\begin{document}
\maketitle
\begin{abstract}

Physical systems are commonly represented as a combination of particles, the individual dynamics of which govern the system dynamics. However, traditional approaches require the knowledge of several abstract quantities such as the energy or force to infer the dynamics of these particles. Here, we present a framework, namely, \textit{Lagrangian graph neural network} (\name), that provides a strong inductive bias to learn the Lagrangian of a particle-based system directly from the trajectory. 
We test our approach on challenging systems with constraints and drag---\name outperforms baselines such as feed-forward $Lagrangian$ neural network (\lnn) with improved performance. We also show the \textit{zero-shot} generalizability of the system by simulating systems two orders of magnitude larger than the trained one and also hybrid systems that are unseen by the model, a unique feature. The graph architecture of \name significantly simplifies the learning in comparison to \lnn with $\sim$25 times better performance on $\sim$20 times smaller amounts of data. Finally, we show the interpretability of \name, which directly provides physical insights on drag and constraint forces learned by the model. \name can thus provide a fillip toward understanding the dynamics of physical systems purely from observable quantities. 
\end{abstract}
\section*{Introduction}
Modeling physical systems involves solving the differential equations governing their dynamics~\cite{lavalle2006planning}. These equations, in turn, assumes the knowledge on the functional form of abstract quantities representing the system such as the forces, energy, $Lagrangian$ or $Hamiltonian$ and parameters associated with them~\cite{goldstein2011classical}. Recently, it has been shown that these dynamics can learned directly from the data using data-driven and physics-informed neural networks (PINN)\cite{karniadakis2021physics,greydanus2019hamiltonian,sanchez2019hamiltonian,lnn, cranmer2020discovering}, which significantly simplifies the model development of complex physical systems.

Among PINNs, a particular family of interest is the Lagrangian (\lnns) and Hamiltonian neural networks (\hnns), where a neural network is parameterized to learn the Lagrangian (or Hamiltonian) of the system directly from the trajectory~\cite{lnn,lnn1,lnn2,sanchez2019hamiltonian,greydanus2019hamiltonian,zhong2021benchmarking,roehrl2020modeling}. The dynamics simulated by \lnn and \hnn has been shown to preserve their own energy during roll-out, thereby respecting basic physical laws. Finzi et al.~\cite{lnn1} showed that formulating \lnns in Cartesian coordinates with explicit constraints simplifies the learning process. Similarly, Zhong et al.~\cite{zhong2020dissipative,zhong2021extending,zhong2020unsupervised} extended \hnns to dissipitative systems, contacts, and videos making the simulations more realistic. In \lnns, it is assumed that since both the predicted and trained \lnns exhibit energy conservation, their energy do not diverge during the roll-out. However, it has been shown that the energy violation in \lnns and \hnns grow with the trajectory~\cite{ICLR2022}. Moreover, these approaches have been demonstrated on systems where the number of degrees of freedom are not large~\cite{lnn,lnn1,lnn2,ICLR2022}. The error in energy violation is empirically shown to increase with the system size~\cite{ICLR2022}. Thus, extending these models to realistic systems with large degrees of freedom is posed as a challenging problem. 

An approach that has received lesser attention in physical systems is to learn their dynamics based on the topology~\cite{cranmer2020discovering,sanchez2020learning}. To this extent, graph-based simulators have been used in physical and molecular systems, successfully~\cite{cranmer2020discovering,sanchez2020learning,greydanus2019hamiltonian,park2021accurate,schoenholz2020jax}. Here, the dynamics of a system can be broken down to the dynamics of its individual components, which are then bound together based on their topological constraints. This raises an interesting question: \textit{can we learn the dynamics of physical systems directly from their trajectory and infer it on significantly larger ones, while preserving the desirable physical laws such as energy and momentum conservation?}

To address this challenge, here, we propose a Lagrangian graph neural network (\name), where we represent the physical system as a graph. Further, inspired from physics, we model the kinetic and potential energy at the node and edge levels, which is then aggregated to compute the total Lagrangian of the system. To model realistic systems, we extend the formulation of \lnns to include Pfaffian constraints, dissipative forces such as friction and drag, and external forces---crucial in applications such as robotics, and computer graphics. Our approach significantly simplifies the learning and improves the performance on complex systems and exhibits generalizability to unseen time steps, system sizes, and even hybrid systems.

\section*{Theory and Architecture}
\subsection*{Lagrangian formulation of particle systems}
The Lagrangian formulation presents an elegant framework to predict the dynamics of an entire system, based on a single scalar function known as the Lagrangian $\mathcal{L}$ (see Methods section for details). The general form of Euler-Lagrange (EL) equation considering \textit{Pfaffian} constraints, drag or other dissipative forces, and external force can be written as
\begin{equation}
    \frac{d}{dt}\frac{\partial \mathcal{L}}{\partial \dot{q}}-\frac{\partial \mathcal{L}}{\partial q} + A^T(q)\lambda - \Upsilon -F =0
 \end{equation}
where $M = \frac{\partial}{\partial \dot{q}}\frac{\partial \mathcal{L}}{\partial \dot{q}}$ represents the mass matrix, $C = \frac{\partial}{\partial q}\frac{\partial \mathcal{L}}{\partial \dot{q}}$ represents Coriolis-like forces, $\Pi = \frac{\partial \mathcal{L}}{\partial q}$ represents the conservative forces derivable from a potential, and $A(q)\in \mathbb{R}^{k\times D}$ represents $k$ velocity constraints (see Supplementary material for details). The acceleration of a particle can then be computed as
\begin{equation}
    {\ddot{q}} = M^{-1} \left(\Pi-C\dot{q} + \Upsilon - A^T(AM^{-1}A^T)^{-1}
    \left( AM^{-1}(\Pi-C\dot{q}+\Upsilon+F )+\dot{A}\dot{q} \right) +F \right)
    \label{eq:acc}
\end{equation}
which can be integrated to obtain the updated configurations. Thus, once the \textit{Lagrangian} of a system is learned, the trajectory of a system can be inferred using the Eq.~\ref{eq:acc}.

\begin{figure}
\includegraphics[width=\textwidth]{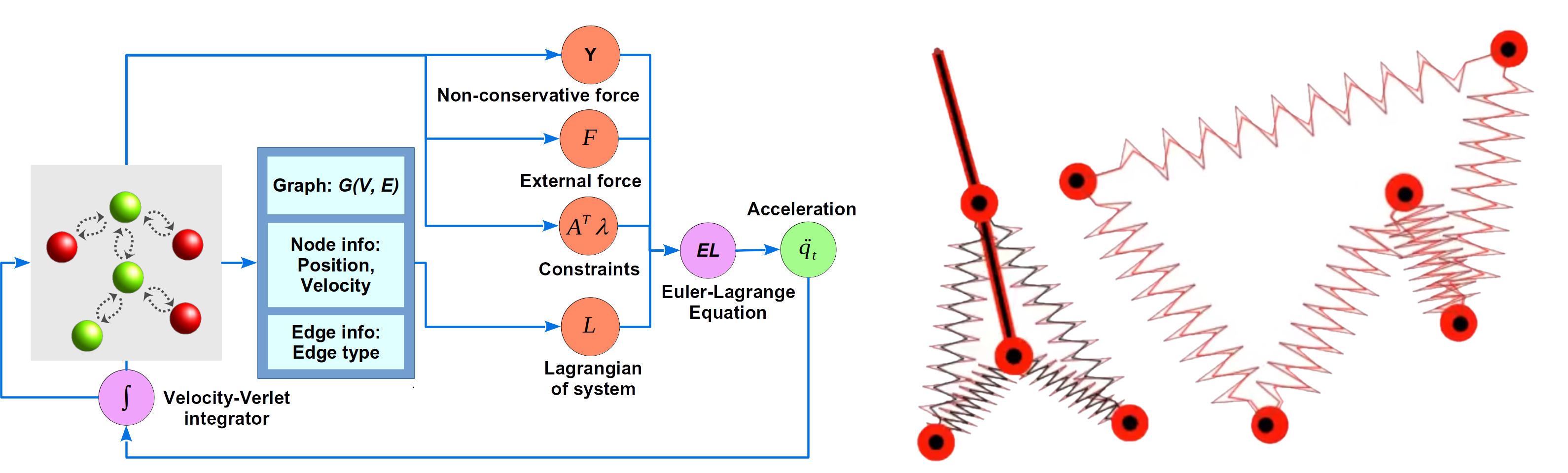}
\caption{The \name framework. Visualization of a hybrid 2-pendulum-4-spring system and a 5-spring system.}
\label{fig:integ}
\end{figure}

In this work, we employ a Lagrangian graph neural network (\name) to learn the Lagrangian of the system directly from its trajectory. An overview of the \name framework, along with example systems, is provided in Fig.~\ref{fig:integ}. The graph topology is used to predict the Lagrangian $\mathcal{L}$, and non-conservative forces such as drag $\Upsilon$. This is combined with the constraints and any external forces in the EL equation to predict the future state. The parameters of the model are learned by minimizing the mean squared error (MSE) over the predicted and true positions across all the particles in the system ( Eq.~\ref{eq:lossfunction}). We next discuss the neural architecture of \name to predict the Lagrangian.

\subsection*{Neural Architecture of \name}
\begin{figure*}
\centering
\includegraphics[width=\linewidth]{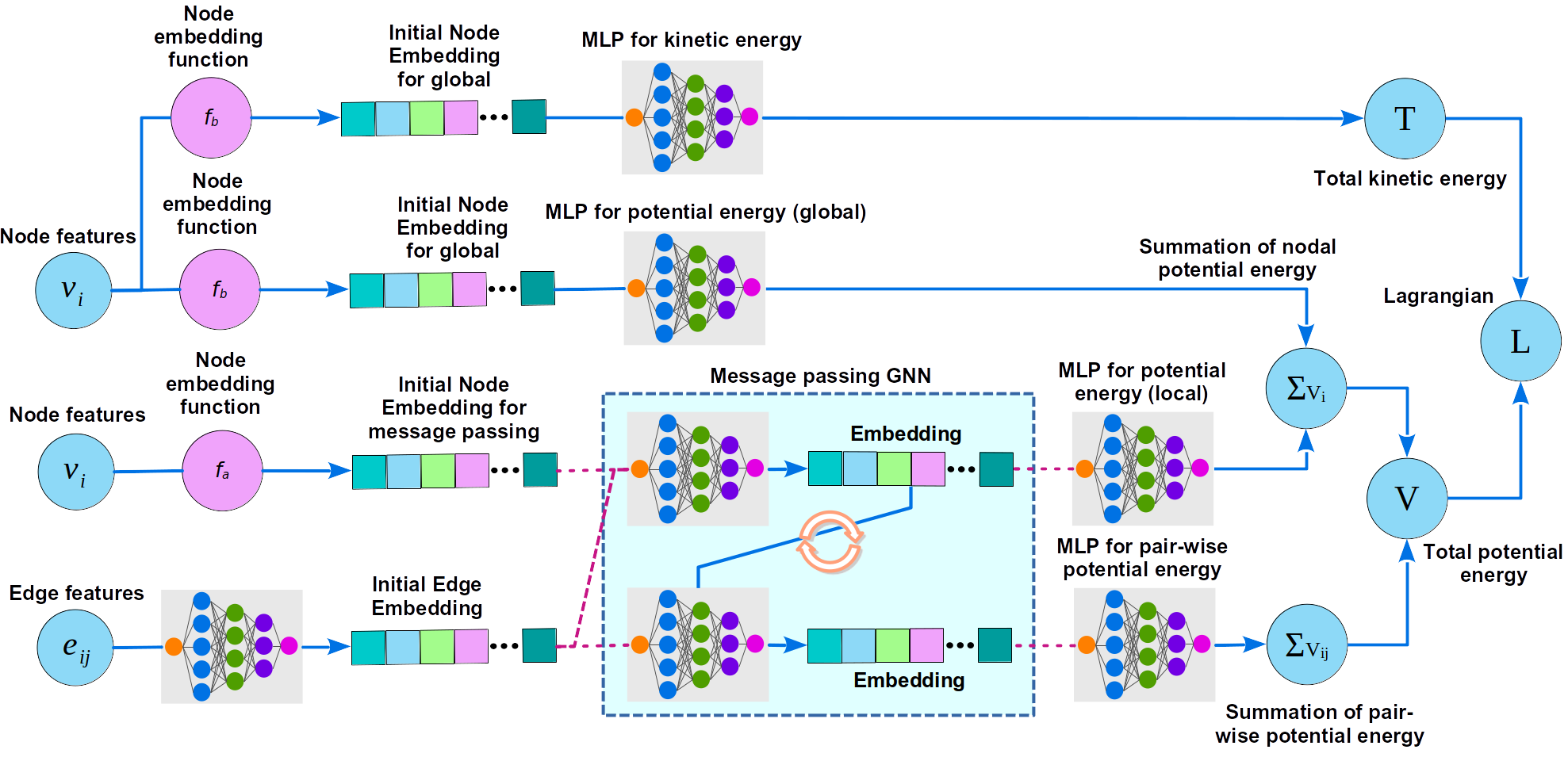}
\caption{General architecture of Lagrangian Graph Neural Network. Note that the drag predictions are similar to the kinetic energy computations. However, the drag is directly featured in $EL$ equation and hence is not shown here.}
\label{fig:architecture}
\end{figure*}
The neural architecture of \name is outlined in Fig.~\ref{fig:architecture}. The aim of the \name is to directly predict the Lagrangian of a physical system while accounting for its topological features. As such, the creation of a graph structure that corresponds to the physical system forms the most important step. To explain the idea better, we will use the running examples of two systems: (i) a double pendulum and (ii) two balls connected by a spring, that represent two distinct cases of systems with rigid and deformable connections.\\
\textbf{Graph structure.} The state of an $n$-body system is represented as a undirected graph, $\mathcal{G}=\{\mathcal{U},\mathcal{E}\}$, where nodes $u_i \in \mathcal{U}$ represents the particles, $\mid\CU\mid=n$, and $e_{ij} \in \mathcal{E}$ represents edges corresponding to constraints or interactions. In our running examples, the nodes correspond to the pendulum bobs and balls, respectively, and the edges correspond to the pendulum rods and springs, respectively. Note that the edges of the graphs may be predefined as in the case of a pendulum or spring system. The edges may also be defined as a function of the distance between two nodes such as $\mathcal{E}=\{e_{ij}=(u_i,u_j)\mid d(u_i,u_j)\leq \theta\}$ where $d(u_i,u_j)$ is a distance function over node positions and $\theta$ is a distance threshold. This allows a dynamically changing edge set based on a cutoff distance as in the case of a realistic breakable spring.\\
\textbf{Input features.} Each node $u_i\in\CU$ is characterized by its type $t_i$, position $q_i=(x_i,y_i,z_i)$, and particle velocity ($\dot{q_i}$). The type $t_i$ is a discrete variable and is useful in distinguishing particles of different characteristics within a system (Ex. two different types of balls). For each edge $e_{ij}$, we store an edge weight $w_{ij}=d(u_i,u_j)=\parallel q_i-q_j\parallel_2$.

Finally, we note that the node input features are classified into two: global and local features. Global features are the ones that are not relevant to the topology and hence are not included in message passing. In contrast, local features are the ones that are involved in message passing. In our running example, while particle type is a local feature, particle position and velocity are global features. \\
\textbf{Pre-Processing:} In the pre-processing layer, we construct a dense vector representation for each node $u_i$ and edge $e_{ij}$ using $\texttt{MLP}_{em}$ as:
\begin{alignat}{2}
    \ch^0_i &= \sqp(\MLP_{em}(\texttt{one-hot}(t_i))) \label{eq:one-hot}\\
    \ch^0_{ij} &= \sqp(\MLP_{em}(e_{ij}))
\end{alignat}
$\sqp$ is an activation function. As the acceleration of the particles is obtained using the $EL$ equation, our architecture requires activation functions that are \textit{doubly-differentiable} and hence, $\sqp$ is an appropriate choice. In our implementation, we use different $\texttt{MLP}_{em}$s for node representation corresponding to kinetic energy, potential energy, and drag. For brevity, we do not separately write the $\texttt{MLP}_{em}$s in Equation~\ref{eq:one-hot}.\\
\textbf{Kinetic energy and drag prediction.} Since the graph uses Cartesian coordinates, the mass matrix is diagonal and the kinetic energy $\tau_i$ of a particle depends only on the velocity $\dot{q}_i$ and mass $m_{i}$ of the particle. Here, we learn the parameterized masses of each particle type based on the embedding $\ch^0_i$. Thus, the $\tau_i$ of a particle is predicted as $\tau_i = \texttt{squareplus}(\texttt{MLP}_{\tau}(\ch^0_i \parallel \dot{q}_i))$, where $\parallel$ represent the concatenation operator, \texttt{MLP}$_{\tau}$ represents an MLP for learning the kinetic energy function, and \texttt{squareplus} represents the activation function. The total kinetic energy of the system is computed as $\mathcal{T}=\sum_{i=1}^{n} \tau_i$. The drag of a particle also typically depends linearly or quadratically to the velocity of the particle. Here, we compute the total drag of a system as $\Upsilon = \sum_i^n \texttt{squareplus}(\texttt{MLP}_{\Upsilon}(\ch^0_i \parallel \dot{q}_i))$. Note that while $\mathcal{T}$ directly goes into the prediction of $\mathcal{L}$, $\Upsilon$ is featured in the $EL$ equation.\\
\textbf{Potential energy prediction.} In many cases, the potential energy of a system is closely dependent on the topology of the structure. To capture this information, we employ multiple layers of \textit{message-passing} between the nodes and edges. In the $l^{th}$ layer of message passing, the node embedding is updated as:
\begin{equation}
    \ch_i^{l+1} = \texttt{squareplus} \left(  \MLP\left(\ch_i^{l}+\sum_{j \in \mathcal{N}_i}\cW_{\CU}^l\cdot\left(\ch_j^l || \ch_{ij}^l\right) \right)\right)
\end{equation}
where, $\mathcal{N}_i=\{u_j \in\CU \mid (u_i,u_j)\in\CE \}$ are the neighbors of $u_i$. $\cW_{\CU}^{l}$ is a layer-specific learnable weight matrix.
$\ch_{ij}^l$ represents the embedding of incoming edge $e_{ij}$ on $u_i$ in the $l^{th}$ layer, which is computed as follows.
\begin{equation}
    \ch_{ij}^{l+1} = \texttt{squareplus} \left( \MLP\left(\ch_{ij}^{l} + \cW_{\CE}^{l}\cdot\left(\ch_i^l || \ch_{j}^l\right)\right) \right)
\end{equation}
Similar to $\cW_{\CU}^{l}$, $\cW_{\CE}^{l}$ is a layer-specific learnable weight matrix specific to the edge set. The message passing is performed over $L$ layers, where $L$ is a hyper-parameter. The final node and edge representations in the $L^{th}$ layer are denoted as $\cz_i=\ch_i^L$ and $\cz_{ij}=\ch_{ij}^L$ respectively.

The total potential energy of an $n$-body system is represented as $\mathcal{V}= \sum_{u_i\in\CU} v_i + \sum_{e_{ij}\in\CE} v_{ij}$, where $v_i$ represents the energy of $u_i$ due to its position and $v_{ij}$ represents the energy due to interaction $e_{ij}$. In our running example, $v_i$ represents the potential energy of a bob in the double pendulum due to its position in a gravitational field, while $v_{ij}$ represents the energy of the spring connected with two particles due to its expansion and contraction. In \name, $v_i$ is predicted as $v_i = \texttt{squareplus}(\texttt{MLP}_{v_i}(\ch_i^0 \parallel q_i))$. Pair-wise interaction energy $v_{ij}$ is predicted as $v_{ij} = \texttt{squareplus} (\texttt{MLP}_{v_{ij}}(\cz_{ij}))$. Although not used in the present case, sometime $v_i$ can be a function of the local features dependent on the topology. For instance, in atomic systems, the charge of an atom can be dependent on neighboring atoms and hence the potential energy due to these charges can be considered as a combination of topology-dependent features along with the global features. In such cases, $v_i$ is predicted as $v_i =\texttt{squareplus}(\texttt{MLP}_{v_i}(\ch_i^0 \parallel q_i)) + \texttt{squareplus}(\texttt{MLP}_{\texttt{mp},v_i}(\cz_i))$, where $\texttt{MLP}_\texttt{mp}$ represents the node $MLP$ associated with the message-passing graph architecture.\\ 
\textbf{Trajectory prediction and training.} Based on the predicted $\mathcal{V}, \mathcal{T}$, and $\Upsilon$, the acceleration $\hat{\ddot{q}}_i$ is computed using the EL Equation \ref{eq:acc}. The loss function of \name is on the predicted and actual accelerations at timesteps $2, 3,\ldots,T$ in a trajectory $\mathbb{T}$, which is then back-propagated to train the MLPs. Specifically, the loss function is as follows.
\begin{equation}
    \label{eq:lossfunction}
  \mathfrak{L}= \frac{1}{n}\left(\sum_{i=1}^n \left(\ddot{q}_i^{\mathbb{T},t}-\left(\hat{\ddot{q}}_i^{\mathbb{T},t}\right)\right)^2\right)
\end{equation}
 Here, $(\hat{\ddot{q}}_i^{\mathbb{T},t})$ is the predicted acceleration for the $i^{th}$ particle in trajectory $\mathbb{T}$ at time $t$ and $\ddot{q}_i^{\mathbb{T},t}$ is the true acceleration. $\mathbb{T}$ denotes a trajectory from $\mathfrak{T}$, the set of training trajectories. It is worth noting that the accelerations are computed directly from the position using the velocity-Verlet update and hence training on the accelerations is equivalent to training on the updated positions.
\section*{Experimental systems}
\label{app:exp_sys}
\subsection*{$n$-Pendulum}
In an $n$-pendulum system, $n$-point masses, representing the bobs, are connected by rigid bars which are not deformable. These bars, thus, impose a distance constraint between two point masses as 
\begin{equation}
||q_{i}-q_{i-1}|| = l_i   
\end{equation}
where, $l_i$ represents the length of the bar connecting the $(i-1)^{th}$ and $i^{th}$ mass. This constraint can be differentiated to write in the form of a $Pfaffian$ constraint as
\begin{equation}
    q_i\dot{q}_i-q_{i-1}\dot{q}_{i-1}=0
\end{equation}
Note that such constraint can be obtained for each of the $n$ masses considered to obtain the $A(q)$.

The Lagrangian of this system has contributions from potential energy due to gravity and kinetic energy. Thus, the Lagrangian can be written as
\begin{equation}
    \mathcal{L}=\sum_{i=1}^n \left(1/2m_i\dot{q_i}^\texttt{T}\dot{q_i}-m_igy_i\right)
\end{equation}
where $g$ represents the acceleration due to gravity in the $y$ direction.
\subsection*{$n$-spring system}
In this system, $n$-point masses are connected by elastic springs that deform linearly with extension or compression. Note that similar to a pendulum setup, each mass $m_i$ is connected only to two masses $m_{i-1}$ and $m_{i+1}$ through springs so that all the masses form a closed connection. The Lagrangian of this system is given by
\begin{equation}
    \mathcal{L}=\sum_{i=1}^n 1/2m_i\dot{q_i}^\texttt{T}\dot{q_i}- \sum_{i=1}^n 1/2k(||q_{i-1}-q_{i}||-r_0)^2
\end{equation}
where $r_0$ and $k$ represent the undeformed length and the stiffness, respectively, of the spring.
\subsection*{Hybrid system}
The hybrid system is a combination of the pendulum and spring system. In this system, a double pendulum is connected to two additional masses through four additional springs as shown in Figure~\ref{fig:integ} with gravity in $y$ direction. The constraints as present in the pendulum system are present in this system as well. The Lagrangian of this four-mass system is \begin{equation}
    \mathcal{L}=\sum_{i=1}^4 1/2m_i\dot{q_i}^\texttt{T}\dot{q_i}- \sum_{i=1}^2 m_igy_i - \sum_{i=3}^4 1/2k(||q_{i-1}-q_{i}||-r_0)^2
\end{equation}
where $m_1$ and $m_2$ represent masses in the double pendulum and $m_3$ and $m_4$ represent masses connected by the springs.
\begin{thm}
\label{thm:momentum}
\textit{In the absence of an external field, \name exactly conserves the momentum of a system.}
\end{thm}
A detailed proof is given in Supplementary Material~\ref{app:mo_cons}. As shown empirically later, this momentum conservation in turn reduces the energy violation error in \name.
\section*{Implementation details}
\label{app:imple}
\subsection*{Dataset generation}

\textbf{Software packages:} numpy-1.20.3, jax-0.2.24, jax-md-0.1.20, jaxlib-0.1.73, jraph-0.0.1.dev0 

\textbf{Hardware:}
Memory: 16GiB System memory,
Processor: Intel(R) Core(TM) i7-10750H CPU @ 2.60GHz

All the datasets are generated using the known Lagrangian of the pendulum and spring systems, along with the constraints, as described in Section~\ref{app:exp_sys}. For each system, we create the training data by performing forward simulations with 100 random initial conditions. For the pendulum and hybrid systems, a timestep of $10^{-5}s$ is used to integrate the equations of motion, while for the spring system, a timestep of $10^{-3}s$ is used. The velocity-verlet algorithm is used to integrate the equations of motion due to its ability to conserve the energy in long trajectory integration. 

All 100 simulations for pendulum and spring system were generated with 100 datapoints per simulation. The was datapoints were collected every 1000 and 100 timesteps for the pendulum and spring systems, respectively. Thus, each training trajectory of the spring and pendulum systems are $10s$ and $1s$ long, respectively. It should be noted that in contrast to the earlier approach, here, we do not train from the trajectory. Rather, we randomly sample different states from the training set to predict the acceleration. For simulating drag, the training data is generated for systems without drag and with linear drag given by $-0.1\dot{q}$, for each particle.\\
\subsection*{Architecture}
For \lnn, we follow a similar architecture suggested in the literature~\cite{lnn,lnn1}. Specifically, we use a fully connected feedforward neural network with 2 hidden layers each having 256 hidden units with a square-plus activation function. For \name, all the MLPs consist of two layers with 5 hidden units, respectively. Thus, \name has significantly lesser parameters than \lnn. The message passing in the \name was performed for two layers in the case of pendulum and one layer in the case of spring.

The kinetic energy term in \lnn is handled as in the earlier case where in the parametrized masses are learned as a diagonal matrix. In the case of \name, the masses are directly learned from the one-hot embedding of the nodes. Interestingly, we observe that the mass matrix exhibits a diagonal nature in \name without enforcing any conditions due to its topology-aware nature.

\subsection*{Training details}
The training dataset is divided in 75:25 ratio randomly, where the 75\% is used for training and 25\% is used as the validation set. Further, the trained models are tested on its ability to predict the correct trajectory, a task it was not trained on. Specifically, the pendulum systems are tested for $10s$, that is $10^6$ timesteps, and spring systems for $20s$, that is $2\times10^4$ timesteps on 100 different trajectories created from random initial conditions. All models are trained for 10000 epochs with early stopping. A learning rate of $10^{-3}$ was used with the Adam optimizer for the training. The performance of both $L^1$ and $L^2$ loss functions were evaluated and $L^2$ was chosen for the final model. The results of \name and \lnn with both the losses are provided in the Supplementary Material.\\

$\bullet$\textbf{Lagrangian Graph Neural Network}
\begin{center}
\begin{tabular}{ |c|c| } 
 \hline
 \textbf{Parameter} & \textbf{Value} \\ 
 \hline
 Node embedding dimension & 5 \\ 
 Edge embedding dimension & 5 \\ 
 Hidden layer neurons (MLP) & 5 \\ 
 Number of hidden layers (MLP) & 2 \\ 
 Activation function & squareplus\\
 Number of layers of message passing (pendulum) & 2\\
 Number of layers of message passing (spring) & 1\\
 Optimizer & ADAM \\
 Learning rate & $1.0e^{-3}$ \\
 Batch size & 100 \\
 \hline
\end{tabular}
\end{center}

$\bullet$\textbf{Lagrangian Neural Network}
\begin{center}
\begin{tabular}{ |c|c| } 
 \hline
 \textbf{Parameter} & \textbf{Value} \\ 
 \hline
 Hidden layer neurons (MLP) & 256 \\ 
 Number of hidden layers (MLP) & 2 \\ 
 Activation function & squareplus\\
 Optimizer & ADAM \\
 Learning rate & $1.0e^{-3}$ \\
 Batch size & 100 \\
 \hline
\end{tabular}
\end{center}

$\bullet$\textbf{Trajectory visualization}
\label{app:video}
For visualization of trajectories of actual and trained models, videos are provided as supplementary material. The supplementary materials contains:\\
(a) \textbf{Hybrid system:} Double pendulum and 2 masses connected with 4 springs under gravitational force as shown in Figure ~\ref{fig:integ}. \\
(b) \textbf{Hybrid system with external force:} Double pendulum and 2 masses connected with 4 springs under gravitational force as shown in Figure ~\ref{fig:integ}. A force of 10 N in x-direction is applied on the second bob of double pendulum. \\
(c) \textbf{5 spring system:} 5 balls connected with 5 springs as shown in Figure ~\ref{fig:integ}.
\section*{Empirical Evaluations}
\section*{\name for $n$-pendulum and $n$-spring systems}
In this section, we evaluate \name and establish that, \textbf{(1)} \name is more accurate than \lnn in modeling physical systems, and \textbf{(2)} owing to topology-aware inductive modeling, \name generalizes to unseen systems, hybrid systems with no deterioration in efficacy for increasing system sizes.
In order to evaluate the performance of \name, we first consider two standard systems, that has been widely studied in the literature, namely, $n$-pendulum and $n$-spring systems~\cite{lnn,lnn1,lnn2,greydanus2019hamiltonian}, with $n=(3,4,5)$. Following the work of~\cite{lnn1}, we evaluate  performance by computing the relative error in \textbf{(1)} the trajectory, known as the \textit{rollout error}, given by $RE(t)=||\hat{q}(t)-q(t)||_2/(||\hat{q}(t)||_2+||q(t)||_2)$ and \textbf{(2)} \textit{energy violation error} given by $||\hat{\mathcal{H}}-\mathcal{H}||_2/(||\hat{\mathcal{H}}||_2+||\mathcal{H}||_2$). The training data is generated for systems without drag and with linear drag given by $-0.1\dot{q}$, for each particle. The timestep used for the forward simulation of the pendulum system is $10^{-5}s$ with the data collected every 1000 timesteps and for the spring system is $10^{-3}s$ with the data collected every 100 timesteps. The details of the experimental systems, data-generation procedure, the hyperparameters and training  procedures are provided in the Methods section.

\name is compared with \textbf{(1)} \lnn in Cartesian coordinates with constraints as proposed by ~\cite{lnn1}, which gives the state-of-the-art performance. Since, Cartesian coordinates are used, the parameterized masses are learned as a diagonal matrix as performed in~\cite{lnn1}. In addition, \name is also compared with two other graph-based approaches suggested in \cite{lnn,sanchez2019hamiltonian}, named as \textbf{(2)} Lagrangian graph network (\textsc{Lgn}), and \textbf{(3)} Graph network simulator (\textsc{Gns}). \textsc{Lgn} exploits a graph architecture to predict the Lagrangian of the physical system, which is then used in the EL equation. \textsc{Gns} uses the position and velocity of the particles to directly predict their updates for a future timestep\cite{greydanus2019hamiltonian}. For both these approaches, since no details of the exact architecture is provided\cite{lnn,sanchez2019hamiltonian}, a full graph network is employed\cite{sanchez2020learning}. 
All the simulations and training were performed in the JAX environment~\cite{schoenholz2020jax}. The graph architecture of \name was developed using the jraph package~\cite{jraph2020github}. All the codes related to dataset generation and training are available as Supplementary Material.\\

\begin{figure}[htp]
\centering
\begin{subfigure}{\textwidth}
\includegraphics[width=\textwidth]{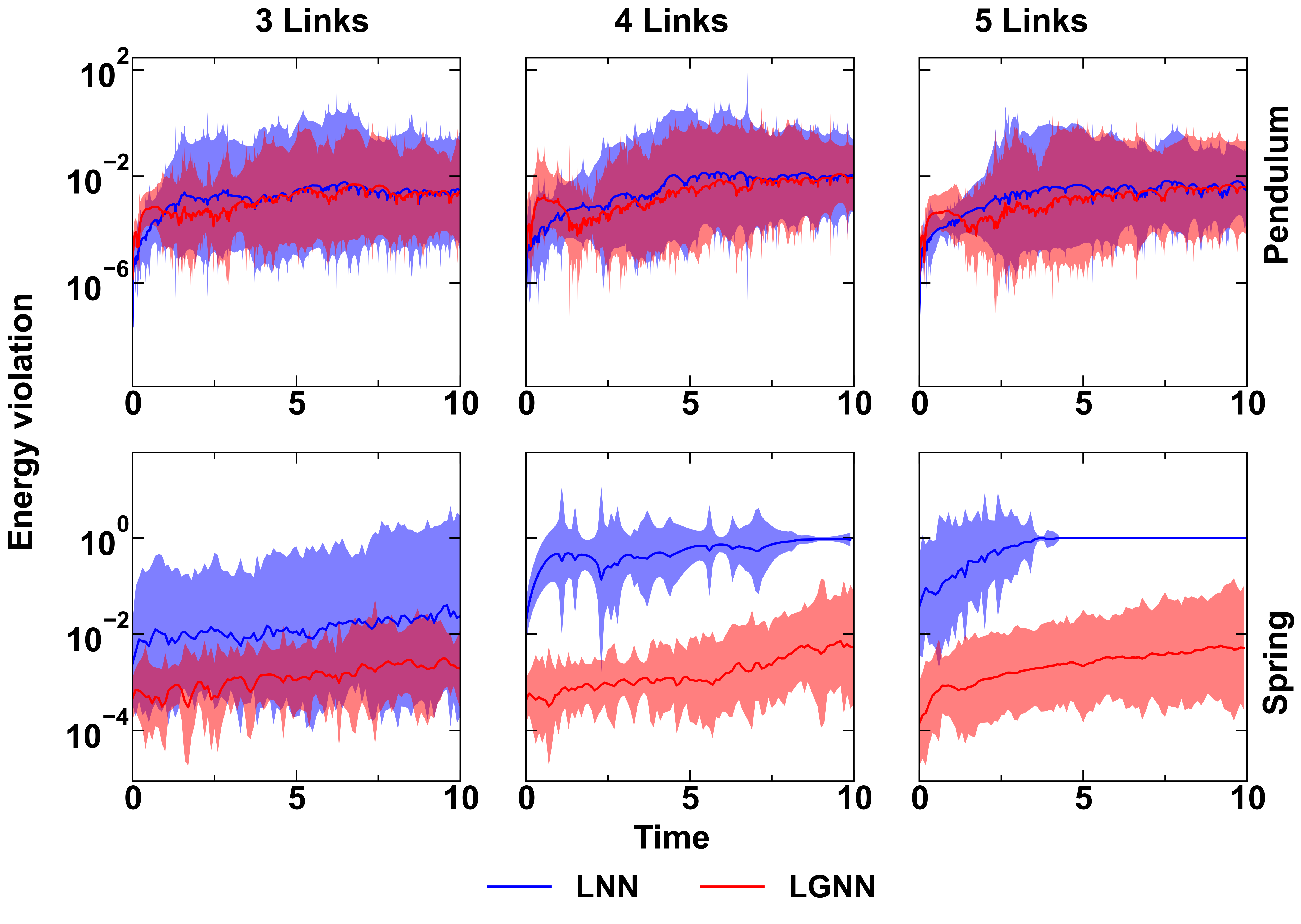}
\end{subfigure}
\begin{subfigure}{\textwidth}
\includegraphics[width=\textwidth]{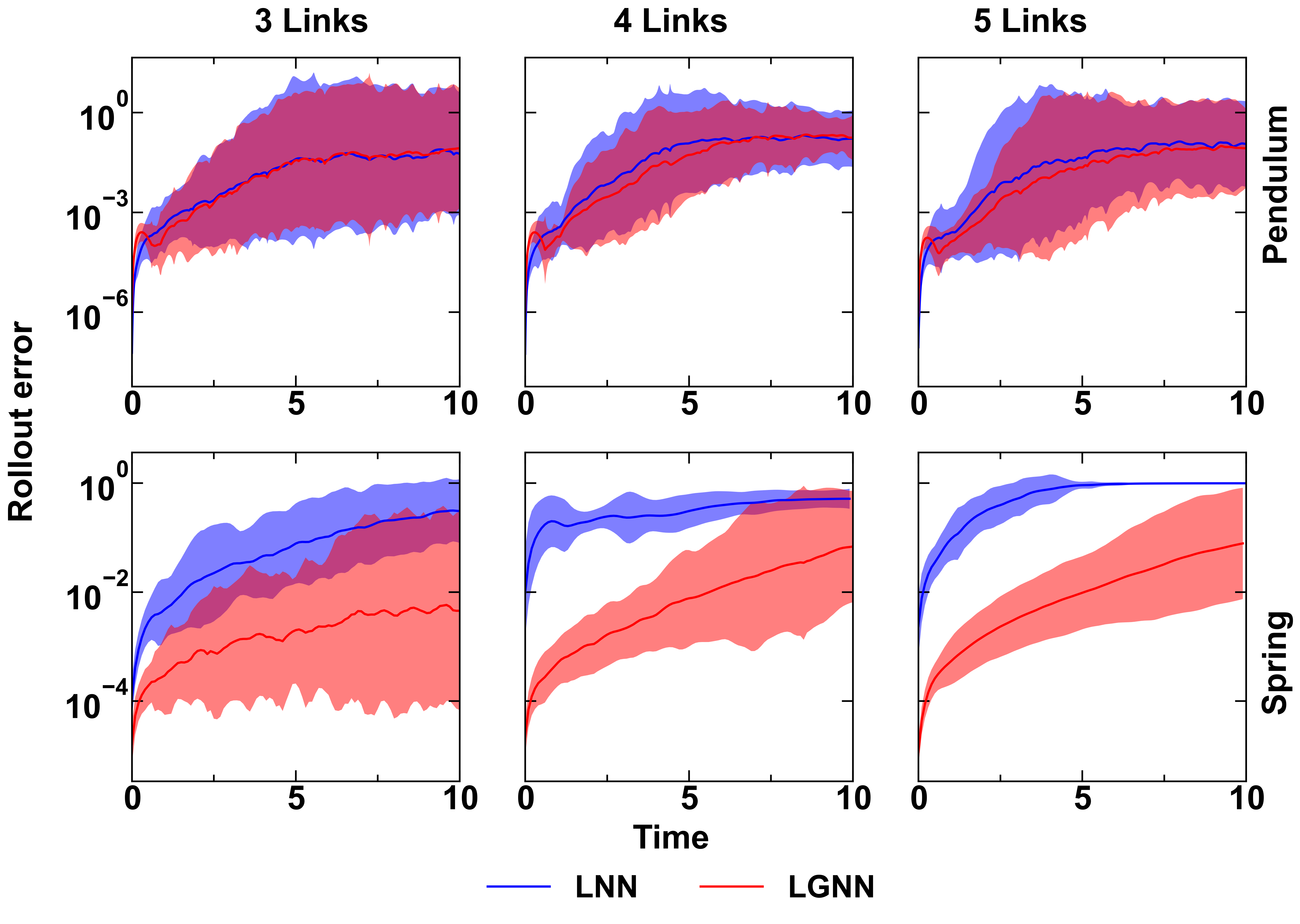}
\end{subfigure}
\caption{Energy violation (left) and rollout error (right) of \name (red) and \lnn (blue) for pendulum and spring systems having 3, 4, and 5 links and subjected to linear drag. Note that the \names for pendulum and springs are trained on 3 and 5 links, respectively, and tested on all the systems. \lnns are trained and tested separately on each of the systems. Shaded region shows the 95\% confidence interval from 100 initial conditions.}
\label{fig:baseline}
\centering
\end{figure}

Figure~\ref{fig:baseline} shows the energy violation and rollout error of \name and \lnn for pendulum and spring systems with 3, 4, and 5 links without drag on each of the particles. For the pendulum system, \name is trained on 3-pendulum and for spring, \name is trained on 5-spring systems. The trained \name is used to perform the forward simulation on the other unseen systems. In contrast, the \lnns are trained and tested on the same systems. We observe that the \name exhibits comparable rollout error with \lnn for pendulum system, even on unseen systems. Further, \name exhibits significantly lower rollout error for spring systems---where the topology plays a crucial role---giving superior performance than \lnns consistently. Due to the chaotic nature of these systems, it is expected that the rollout error diverges. However, energy violation error gives a better evaluation of the realistic nature of the predicted trajectory. Interestingly, we note that the energy violation error is fairly constant for \name for both pendulum and spring systems. Moreover, the error on both seen and unseen systems are comparable suggesting a high degree of generalizability for \name. The geometric mean of energy violation error and rollout error of \lnn and \name for systems with and without drag are provided in the Supplementary material. 

\subsection*{Generalizability}
\label{sec:generalizability}
\begin{figure}[t]
\centering
\begin{subfigure}{\columnwidth}
\includegraphics[width=\columnwidth]{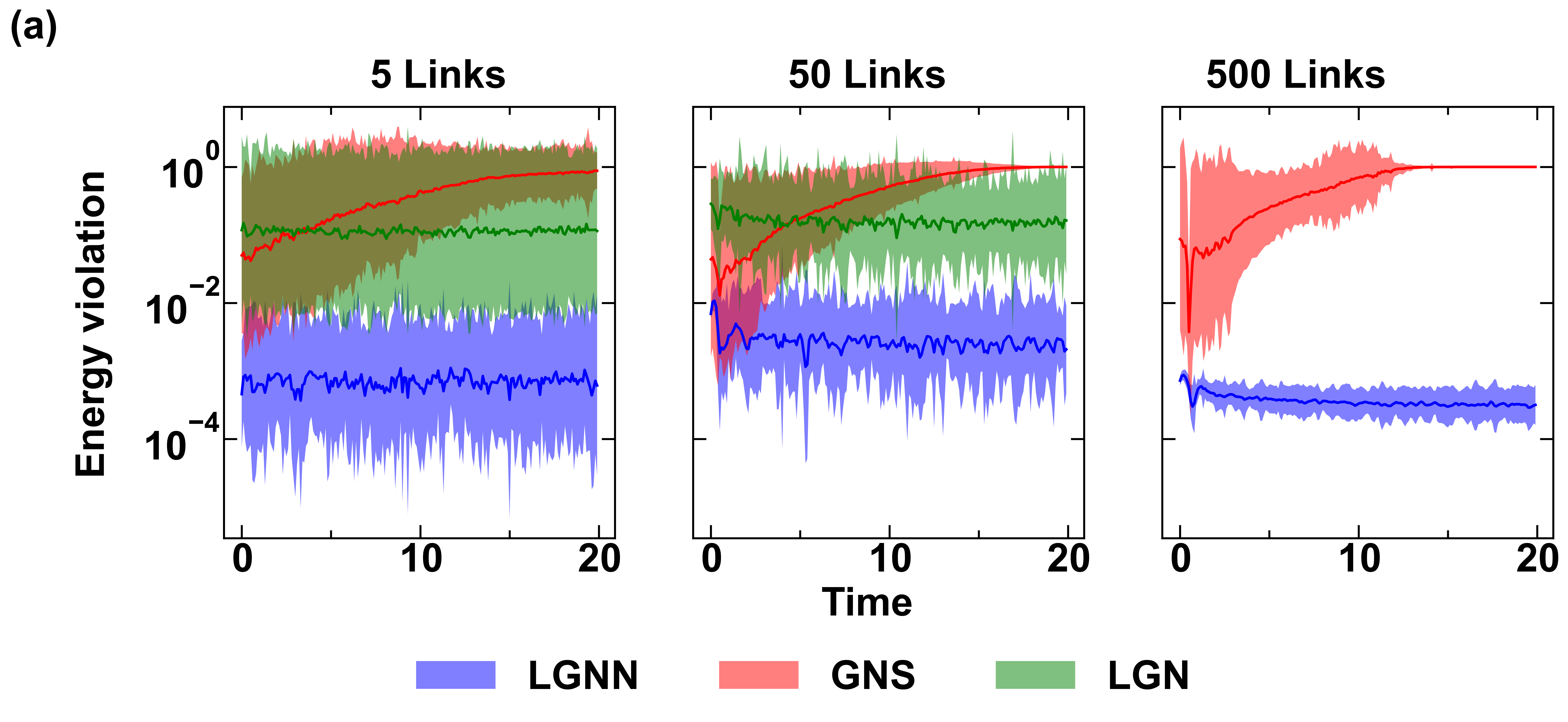}
\end{subfigure}
\begin{subfigure}{\columnwidth}
\includegraphics[width=\columnwidth]{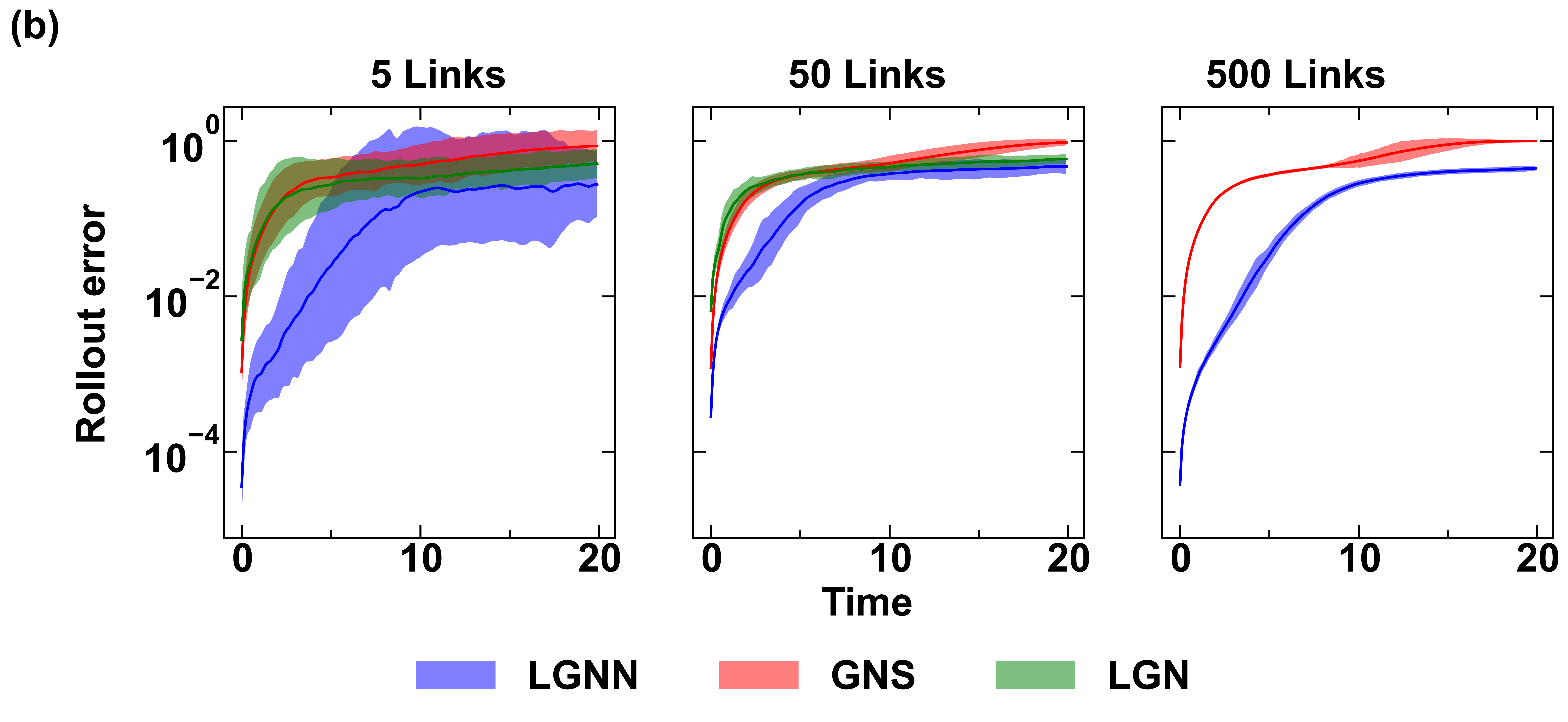}
\end{subfigure}
\caption{(a) Energy violation and (b) rollout error for 5, 50, and 500 spring system with \name trained on 5 spring system. Shaded region shows the 95\% confidence interval from 100 initial conditions.\label{fig:generalizability}}
\end{figure}

Now, we examine the ability of \name to push the limits of the current approaches. A regular \lnn uses a feed-forward NN to take the entire ${q,\dot{q}}$ as the input and then predicts the Lagrangian. Owing to this design, the number of model parameters grows with the input size. Hence, an \lnn trained on an $n$-sized system cannot be applied for inference on an $n'$-sized system, where $n\neq n'$. This makes \lnn \textit{transductive} in nature. In contrast, \name is \textit{inductive}, wherein the parameter space remains constant with system size. This is a natural consequence of our design where the learning happens at node and edge levels. Hence, given an arbitrary-sized system, we only need to predict the potential and kinetic energies at each node and edge, which we then aggregate to predict the combined energies at a system level. Further, a complex system that is unseen, if divisible into multiple sub-graphs with each sub-graph representing a learned system, the overall system can be simulated with zero-shot generalizability.

Figure~\ref{fig:generalizability} shows the energy and rollout error of spring systems with 5, 50, 500 links using \name, \textsc{Lgn} and \textsc{Gns} trained on 5-spring system. We observe that \name performs significantly better than the other baselines. Note that the results of \textsc{Lgn} for 500 springs is not included due to significantly increased computational time for the forward simulation in comparison to the other models. In addition to the rollout error, we observe that the energy violation error of \name remains constant with low values even for systems that are 2 orders of magnitude larger than the training set. This suggests that the \name can produce realistic trajectories on large scale systems when trained on much smaller systems. These results also establish the superior nature of \name architecture, which exhibits improved performance with higher computational efficiency. The performance of similar systems with drag is provided in the Supplementary Material.
\begin{figure}
\centering
\begin{subfigure}{0.46\columnwidth}
\includegraphics[width=\columnwidth]{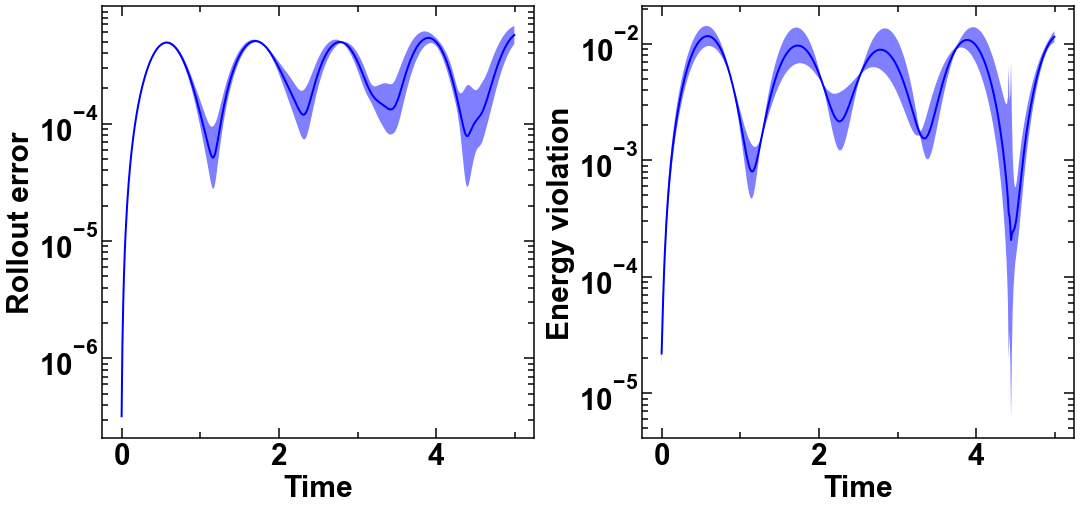}
\end{subfigure}
\begin{subfigure}{0.24\columnwidth}
    \includegraphics[width=\columnwidth]{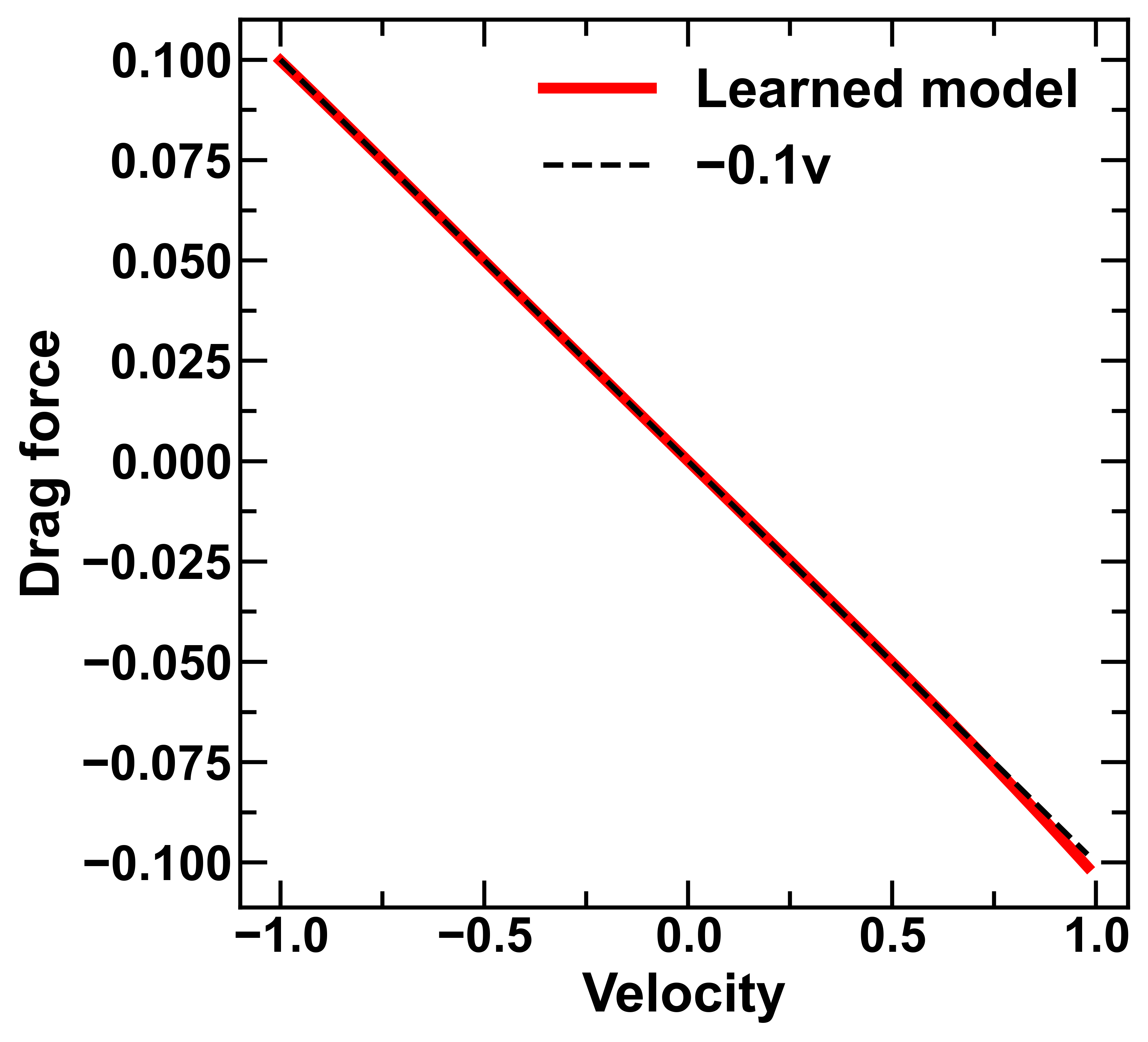}
\end{subfigure}
\begin{subfigure}{0.24\columnwidth}
\includegraphics[width=\columnwidth]{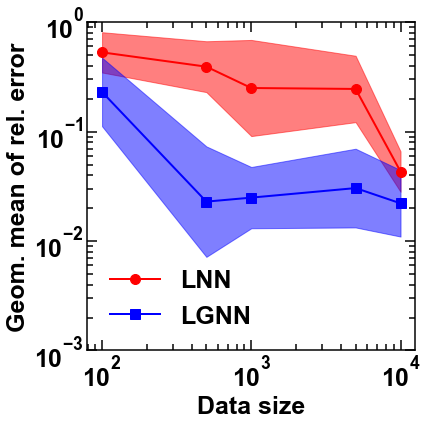}
\end{subfigure}
\caption{(a) Energy violation and (b) rollout error of \name in the unseen hybrid system. (c) Learned node-level drag force. (d) Data efficiency of \lnn and \name on 3-spring system.}
\label{fig:dataeff}
\end{figure}

To test generalizability to an unseen system, we consider a hybrid system with 2-pendulum-4-springs as shown in Figure~\ref{fig:integ}. The \names trained on the 5-spring and 3-pendulum systems are used to model the hybrid system. Specifically, the $\mathcal{V}$ of the physical system is considered as the union of the $\mathcal{V}$ of two subgraphs, one representing the pendulum system and the other representing the spring systems. Then, the computed $\mathcal{V}$ is substituted in the EL equation to simulate the forward trajectory. Figure~\ref{fig:dataeff}(a) and (b) shows the energy violation and rollout error on the hybrid system. We observe that both these errors are quite low and are not growing with time. This suggests that \name, when trained individually on several subsystems, can be combined to form a powerful graph-based simulator. Videos of simulated trajectories under different conditions, such as external force, are provided as supplementary material (see~\ref{app:video}) for visualization.
\subsection*{Interpretability and data efficiency}
\label{sec:interpretability}

The trained \name provides interpretability, thanks to the graph structure. That is, since the computations of \name are carried out at the node and edge levels, the functions learned at these level can be used to gain insight into the behavior of the system. Further, the parameterized masses of the systems are learned directly during the training of \name, leading to a diagonal mass matrix. This is in contrast to \lnn where the prior information that mass matrix remains diagonal in the Cartesian coordinates is imposed to make the learning simpler~\cite{lnn1}. 

Figure~\ref{fig:dataeff}(c) shows the learned drag force as a function of velocity. We observe that the neural network has learned the drag in excellent agreement with the ground truth. Thus, in cases of systems where the ground truth is unknown, the learned drag can provide insights into the nature of dissipative forces in the system. Similarly, the $\lambda$ learned during the training directly provides insights into the magnitude of the constraint forces.

Finally, we analyze the efficiency of \name to learn from the training data. To this extent, we train both the models on a 3-spring system with ($100, 500, 1000, 5000, 10000$) datapoints. Figure~\ref{fig:dataeff}(d) shows the geometric mean of relative error on the trajectory of the system. We observe that \name consistently outperforms \lnn, Further, the performance of \name saturates with 500 datapoints. In addition, \name trained on these 500 datapoints exhibits $\sim$25 and $\sim$3 times better performance in terms of geometric relative error than \lnn trained on $500$ and $10000$ datapoints. This confirms that the topology-aware \name can learn efficiently from small amounts of data, still yield better performance than \lnn.
\section*{Conclusions and outlook}
\label{sec:conclusion}
Altogether, we demonstrate a novel graph architecture, namely \name, which can learn the dynamics of systems directly from the trajectory, while exhibiting superior energy and momentum conservation. We show that the \name can generalize to any system size ranging orders of magnitude, when learned on a small system. Further, \name can even simulate unseen hybrid systems when trained on the systems independently. Finally, we also show that \name exhibits better data-efficiency, learning from smaller amounts of data in comparison to the original \lnn.

There are several future directions that the present work opens up. The \name, at present, is limited to systems without collisions and other deformations. Extending \name to handle elastic and plastic deformations in addition to collisions can significantly widen the application areas. Further, \name assumes the knowledge of constraints. Automating the learning of the constraints while training the model using neural networks can provide a new paradigm to learn the constraints along with the Lagrangian. The graph nature of \name can potentially supplement the learning of constraints. Although, the message passing in \name is translationally and rotationally invariant, the expressive power is limited as only the distance is given as the edge feature. Enhancing the feature representation, and message passing through architectural modifications as in the case of equivariant \textsc{Gnn}s can increase the applicability to more complex applications such as atomistic simulations of materials.

\section*{Data and code availability}
All the codes related to dataset generation and training are made are available as Supplementary Material.

\section*{Competing interests declaration}
Authors declare no competing interests. 

\bibliographystyle{unsrt}
\bibliography{example_paper}

\newpage

\appendix
\section*{\centering \Large{SUPPLEMENTARY MATERIAL}}
\subsection*{Preliminaries on Learning Dynamical Systems \label{sec:prelim}}
The time evolution or $dynamics$ of the state ($q(t),\dot{q}(t)$) of a physical system can be represented as $\ddot{q}=F(q,\dot{q},t)$, where $q, \dot{q}\in \mathbb{R}^D$ is a function of time ($t$) for a system with $D$ degrees of freedom. The future states or $trajectory$ of the system can be predicted by integrating these equations to obtain $q(t+1)$ and so on. While there are several physics-based methods for generating the $dynamics$ of the system such as d'Alembert's principle, Newtonian, Lagrangian, or Hamiltonian approaches, all these approaches result in the equivalent sets of equations~\cite{murray2017mathematical}.

The Lagrangian formulation presents an elegant framework to predict the dynamics of an entire system, based on a single scalar function known as the Lagrangian $\mathcal{L}$. The standard form of Lagrange's equation for a system with $holonomic$ constraints is given by
$\frac{d}{dt} \left( \frac{\partial \mathcal{L}}{\partial \dot{q_i}}\right)-\left( \frac{\partial \mathcal{L}}{\partial q_i} \right) = 0$,
where, $i=1,2,3,\ldots,D$, and the Lagrangian is $\mathcal{L}(q,\dot{q},t)=\mathcal{T}(q,\dot{q},t)-\mathcal{V}(q,t)$ with $\mathcal{T}(q,\dot{q},t)$ and $\mathcal{V}(q,t)$ representing the total kinetic energy of the system and the potential function from which generalized forces can be derived. Accordingly, the dynamics of the system can be represented using \textit{Euler-Lagrange (EL)} equations as ${\ddot{q}_i} = \left( \nabla_{{\dot{q}_i} {\dot{q}_i}} \mathcal{L} \right)^{-1} \left[ \nabla_{{q}_i} \mathcal{L} - \left( \nabla_{{\dot{q}_i} {{q}_i}} \mathcal{L} \right) {\dot{q}_i}\right]$. 

In \lnn, the Lagrangian of a system is learned using a neural network by training on the acceleration of the system. Once trained, the \lnn, along with EL equations, provide the forward trajectory of the system. \lnns are highly desirable for modeling physical systems due to their energy-conserving nature, a consequence of modeling acceleration directly from the EL equations. However, as we demonstrate later, while \lnns conserve the energy of their own trajectory, it is not guaranteed that the predicted energy and ground truth are comparable, even for a well trained \lnn.

\subsection*{Lagrangian mechanics with Pfaffian Constraints, and Dissipation}
The \lnn formulation presented in Section \ref{sec:prelim}, although powerful, is restricted to systems: (i) without any external forces, (ii) with generalized forces that are derivable from a potential function, (iii) without any dissipation such as friction or drag, and (iv) having holonomic constraints. To address these limitations, we extend the formulation of \lnns as detailed below~\cite{murray2017mathematical,lavalle2006planning}.\\
\textbf{Pfaffian Constraints.} Holonomic constraints restrict the allowable configurations of a system to a smooth hyper-surface that satisfies the constraints. The trajectory of the system is governed by the algebraic constraints in terms of the positions and are enforced by the constraint forces associated with each constraint. For example, the movement of the bob in a 2D-pendulum is restricted by its rod following the equation $(x^2-{y}^2)=l^2$, where $(x,y)$ represents the coordinate of the bob with the pin at the origin and $l$ is the length of the bar. Here, the constraint is enforced by the forces in the bar and represents a holonomic constraint. However, in some cases the constraints may depend not on the position, but on the velocities of system. For example, in multi-fingered grasping, the velocity of two or more fingers are constrained so that the combined geometry formed is able to catch or hold an object. Such constraints, known as \textit{Pfaffian constraints}, can be expressed as 
\begin{equation}
    A(q)\dot{q} = 0
    \label{eq:constraint}
\end{equation}
where, $A(q)\in \mathbb{R}^{k\times D}$ represents $k$ velocity constraints. Note that a holonomic constraint can also be written as a Pfaffian constraint. For instance, the holonomic constraint of a pendulum can be written in the Pfaffian form as $x\dot{x}+y\dot{y}=0$. More generally, an integrable Pfaffian constraint is a holonomic constraint, while a non-integrable one is a non-holonomic constraint.\\
\textbf{Dissipation and external forces.} Physical systems are constantly subjected to dissipative forces, $\Upsilon$ such as drag, and friction. For instance, a system through a fluid such as air of water may be subjected to linear ($c_1\dot{q}$, Stoke's law) or quadratic ($c_2\dot{q}^2$, air drag) drag forces resulting in energy dissipation. Similarly, a system may be additionally subjected to external forces $F$ at different degrees of freedom due to interactions with the environment.\\
\textbf{Modified Euler-Lagrange Equation.} Considering the additional forces mentioned above, the modified EL equation can be written as:
\begin{equation}
    \frac{d}{dt}\frac{\partial \mathcal{L}}{\partial \dot{q}}-\frac{\partial \mathcal{L}}{\partial q} + A^T(q)\lambda - \Upsilon -F =0
\end{equation}
where $A^T$ forms a non-normalized basis for the constraint forces, $\lambda \in \mathbb{R}^k$, known as the Lagrange multipliers, gives the relative magnitudes of these force constraints, $\Upsilon$ represents the non-conservative forces, such as friction, which are not directly derivable from a potential, and $F$ represents any external forces acting on the system. This equation can be modified to obtain $\ddot{q}$ as:
\begin{equation}
    {\ddot{q}} = M^{-1}\left(-C \dot{q} +\Pi + \Upsilon - A^T(q)\lambda +F\right)
    \label{eq:EL}
\end{equation}
where $M = \frac{\partial}{\partial \dot{q}}\frac{\partial \mathcal{L}}{\partial \dot{q}}$ represents the mass matrix, $C = \frac{\partial}{\partial q}\frac{\partial \mathcal{L}}{\partial \dot{q}}$ represents Coriolis-like forces, and $\Pi = \frac{\partial \mathcal{L}}{\partial q}$ represents the conservative forces derivable from a potential. Differentiating the constraint Equation (\ref{eq:constraint}) gives
$A(q)\ddot{q}+\dot{A}(q)\dot{q} = 0$. Solving for $\lambda$ and substituting in Equation \ref{eq:EL}, we obtain $\ddot{q}$ as 
\begin{equation}
    {\ddot{q}} = M^{-1} \left(\Pi-C\dot{q} + \Upsilon - A^T(AM^{-1}A^T)^{-1}
    \left( AM^{-1}(\Pi-C\dot{q}+\Upsilon+F )+\dot{A}\dot{q} \right) +F \right)
    \label{eq:acc}
\end{equation}
Thus, for a system subjected to these additional forces, \lnn can be trained by minimizing the loss on the predicted and observed trajectory, where the predicted acceleration $\Hat{\ddot{q}}$ is obtained using the Equation \ref{eq:acc}. It is worth noting that in this equation, $M, C,$ and $\Pi$ can be directly derived from the $\mathcal{L}$. Constraints on the systems, drag, and external forces are generally known. Additionally, we demonstrate later that the drag forces can also be learned using this framework if they are unknown. For particle systems in Cartesian coordinates, the mass matrix $M(q)$ remains constant with only diagonal entries $m_{ii}$. Inducing this as a prior knowledge, wherein the masses are parameterized as a diagonal matrix is shown to simplify the learning process~\cite{lnn1}.

\section{Energy violation of \lnn}
\label{app:ener_vio}
Let the error in dynamics of the \lnn be $e(q) = |\hat{\ddot{q}}_t-\ddot{q}_t|$ be bounded such that $|e(q)|< n \textrm{sup}|\delta_i|$, where $n$ represents the number of particles in the system and $\textrm{sup}|\delta_i|$ is the supremum of error in the dynamics of all particles in the system. To obtain the evolution of energy violation $\varepsilon_t$, we compute
\begin{align}
    \frac{d\varepsilon_t}{dt}&=\frac{d}{dt} \left({\mathcal{H}}(\hat{q}_t,\hat{\dot{q}}_t))-\mathcal{H}(q_t,\dot{q}_t) \right) \nonumber \\
    &= \frac{d{\mathcal{H}}(\hat{q}_t,\hat{\dot{q}}_t))}{dt}-\frac{d\mathcal{H}(q_t,\dot{q}_t)}{dt}=\frac{d{\mathcal{H}}(\hat{q}_t,\hat{\dot{q}}_t))}{dt}; \quad \textrm{since, } \frac{d\mathcal{H}(q_t,\dot{q}_t)}{dt} = 0 \nonumber \\
    \frac{d\varepsilon_t}{dt} &=\frac{d}{dt} \left(\mathcal{L}(\hat{q}_t,\hat{\dot{q}}_t) - \frac{\partial \mathcal{L}(\hat{q}_t,\hat{\dot{q}}_t)}{\partial \hat{\dot{q}}_t} \hat{\dot{q}}_t\right); \textrm{writing $\mathcal{H}$ in terms of $\mathcal{L}$}\nonumber \\
    &= \frac{d\mathcal{L}(\hat{q}_t,\hat{\dot{q}}_t)}{dt} - \frac{d}{dt}\left( \frac{\partial \mathcal{L}(\hat{q}_t,\hat{\dot{q}}_t)}{\partial \hat{\dot{q}}_t}\right)\hat{\dot{q}}_t - \frac{\partial \mathcal{L}(\hat{q}_t,\hat{\dot{q}}_t)}{\partial \hat{\dot{q}}_t}\hat{\ddot{q}}_t \nonumber
\end{align}
Replacing $\frac{d}{dt}=\frac{\partial }{\partial q}\dot{q}+\frac{\partial }{\partial \dot{q}}\ddot{q}$, we get
\begin{align}
    \frac{d\varepsilon_t}{dt} &= \frac{\partial \mathcal{L}(\hat{q}_t,\hat{\dot{q}}_t)}{\partial \hat{q}_t} \hat{\dot{q}}_t + \frac{\partial \mathcal{L}(\hat{q}_t,\hat{\dot{q}}_t)}{\partial \hat{\dot{q}}_t} \hat{\ddot{q}}_t - \frac{d}{dt} \left( \frac{\partial \mathcal{L}(\hat{q}_t,\hat{\dot{q}}_t)}{\partial \hat{\dot{q}}_t} \right) \hat{\dot{q}}_t - \frac{\partial \mathcal{L}(\hat{q}_t,\hat{\dot{q}}_t)}{\partial \hat{\dot{q}}_t}\hat{\ddot{q}}_t \nonumber \\
    &=\left( \frac{\partial \mathcal{L}(\hat{q}_t,\hat{\dot{q}}_t)}{\partial \hat{q}_t} - \frac{d}{dt} \left( \frac{\partial \mathcal{L}(\hat{q}_t,\hat{\dot{q}}_t)}{\partial \hat{\dot{q}}_t} \right) \right)\hat{\dot{q}}_t
\end{align}
The bracketed term represents the error in the EL equation or the rate of change of momentum of the predicted trajectory computed based on the true Lagrangian. If the predicted Lagrangian $\hat{\mathcal{L}}$ and true Lagrangian $\mathcal{L}$ are close enough, then this error should not diverge. However, even for a well-trained \lnn there is no guarantee that this error is bounded, which agrees with the empirical experiments as well. Further, even if this error is bounded, the energy violation grows with the number of particles and the displacement as demonstrated in Theorem \ref{thm:ener_vio}.

\section{Momentum conservation of \name}
\label{app:mo_cons}
Consider systems where the energy is a function of the internal interactions only, for instance, $n$-spring systems or $n$-pendulum in space without gravity. For such systems, the predicted Lagragian $\hat{\mathcal{L}}$ in \name exactly conserves the momentum. In other words, the rate of change of momentum as computed by the predicted $\hat{\mathcal{L}}$ on the real trajectory is zero, that is, $\sum_{i=1}^n \frac{d}{dt} \left( \frac{\partial \hat{\mathcal{L}}({q_{i,t}},{\dot{q}_{i,t}})}{\partial {\dot{q}_{i,t}}} \right)=0$. Note that here we use the real trajectory ($q,\dot{q}$) and predicted Lagrangian ($\hat{\mathcal{L}}$) based on the following argument. We know that ($q,\dot{q}$) presents a momentum conserving trajectory based on the true $\mathcal{L}.$ Thus, if ($\hat{\mathcal{L}}$) and (${\mathcal{L}}$) are equivalent, the predicted trajectory should also be momentum conserving. Conversely, if the time-evolution of $\hat{\mathcal{L}}$ produces a momentum conserving trajectory and $\hat{\mathcal{L}}$ and $\mathcal{L}$ are equivalent, then the ($\hat{\mathcal{L}}$) should conserve the momentum of the true trajectory as well. 
 
To this extent, the error $\sum_{i=1}^n \hat{\delta}_{i,t}$ in the rate of change of momentum as computed using the predicted Lagrangian $\hat{L}$ on the real trajectory $(q,\dot{q})$, is given by 
\begin{align}
    \sum_{i=1}^n \hat{\delta}_{i,t} = \sum_{i=1}^n \left( \frac{\partial \hat{\mathcal{L}}({q}_{i,t},{\dot{q}}_{i,t})}{\partial {q}_{i,t}} - \frac{d}{dt} \left( \frac{\partial \hat{\mathcal{L}}({q}_{i,t},{\dot{q}}_{i,t})}{\partial {\dot{q}}_{i,t}} \right) \right)
\end{align}
Where, $\hat{\mathcal{L}}({q}_{i,t},{\dot{q}}_{i,t})=\sum_{i=1}^{n} \tau_i - (\sum_{i=1}^{n} v_i + \sum_{e_{ij}\in \mathcal{E}} v_{ij})$. In the absence of an external field, the potential energy is a function of only the internal interaction as represented by the edges. Thus, the nodal potential energy $v_i$, which depends on the actual position, goes to zero. Hence, $\hat{\mathcal{L}}({q}_{i,t},{\dot{q}}_{i,t})=\sum_{i=1}^{n} \tau_i - \sum_{e_{ij}\in \mathcal{E}} v_{ij}$. Further, the potential energy $v_{ij}$ of edges is a function of the edge features, ($|q_i-q_j|$), that is, the edge distance, and the kinetic energy $\tau_i$ (in Cartesian coordinates) is a function of $\dot{q_i}$. If the momentum of the system is exactly conserved, $\sum_{i=1}^n \hat{\delta}_{i,t} = 0$. To prove this, it is enough to prove that, $\sum_{i=1}^n \frac{d}{dt} \left( \frac{\partial \hat{\mathcal{L}}({q}_{i,t},{\dot{q}}_{i,t})}{\partial {\dot{q}}_{i,t}} \right) = \sum_{i=1}^n \frac{\partial \hat{\mathcal{L}}({q}_{i,t},{\dot{q}}_{i,t})}{\partial {q}_{i,t}} =0$.

\begin{align}
 \sum_{i=1}^n \frac{\partial \hat{\mathcal{L}}({q}_{i,t},{\dot{q}}_{i,t})}{\partial {q}_{i,t}} &= -\sum_{e_{ij}\in \mathcal{E}}\frac{\partial v_{ij}(|q_i-q_j|)}{\partial q_i} \nonumber \\
 &= -\frac{1}{2} \sum_{e_{ij}\in \mathcal{E}} \left( \frac{\partial v_{ij}(|q_i-q_j|)}{\partial q_i}+\frac{\partial v_{ji}(|q_j-q_i|)}{\partial q_j} \right)
\end{align}

Consider, 
\begin{align}
 -\frac{\partial v_{ij}(|q_i-q_j|)}{\partial q_i}&= -\frac{d v_{ij}(|q_i-q_j|)}{d(|q_i-q_j|)}\frac{\partial  |(q_i-q_j|)}{\partial q_i} \nonumber \\
 &= -\frac{d v_{ij}(|q_i-q_j|)}{d(|q_i-q_j|)}\frac{\partial  (|q_i-q_j|)}{\partial q_i} \nonumber \\
 &= -\frac{d v_{ij}(|q_i-q_j|)}{d(|q_i-q_j|)} \frac{(q_i-q_j)}{|(q_i-q_j)|}  \nonumber \\
 &= -\frac{d v_{ij}(|q_i-q_j|)}{d(|q_i-q_j|)} \left( -\frac{(q_j-q_i)}{|(q_i-q_j)|}\right)  \nonumber \\
 &= \frac{d v_{ji}(|q_j-q_i|)}{d(|q_j-q_i|)} \frac{(q_j-q_i)}{|(q_j-q_i)|}; \nonumber \\
 &\textrm{since, } |q_j-q_i| = |q_i-q_j| \textrm{ and } v_{ij} = v_{ji} \nonumber \\
 &= \frac{d v_{ji}(|q_j-q_i|)}{d(|q_j-q_i|)} \frac{\partial(|q_j-q_i|)}{\partial q_j} \nonumber \\
 &= \frac{\partial v_{ji}(|q_j-q_i|)}{\partial q_j} \nonumber \\
 -\frac{\partial v_{ij}(|q_i-q_j|)}{\partial q_i} -\frac{\partial v_{ji}(|q_j-q_i|)}{\partial q_j} &= 0
\end{align}

Combining, Eqs. (16) and (17),
\begin{align}
    \sum_{i=1}^n \frac{d}{dt} \left( \frac{\partial \hat{\mathcal{L}}({q}_{i,t},{\dot{q}}_{i,t})}{\partial {\dot{q}}_{i,t}} \right) = \sum_{i=1}^n \frac{\partial \hat{\mathcal{L}}({q}_{i,t},{\dot{q}}_{i,t})}{\partial {q}_{i,t}} =0
\end{align}
Thus, we show that \name conserves the momentum exactly on the true trajectory as well. This, in turn, suggests that the error in the energy violation for \name should be lower than that of \lnn.

\section*{Comparison of \name with baselines}
\label{app:baseline_error}
To compare \name with \lnn, we evaluate the energy violation error and rollout error for both systems with and without drag. In addition, we also evaluate the effect of $L^1$ vs $L^2$ loss functions. We observe that the $L^2$ loss provides better performance for all the models.
\begin{figure}[ht]
\begin{subfigure}{\columnwidth}
    \centering
    \includegraphics[width=0.9\columnwidth]{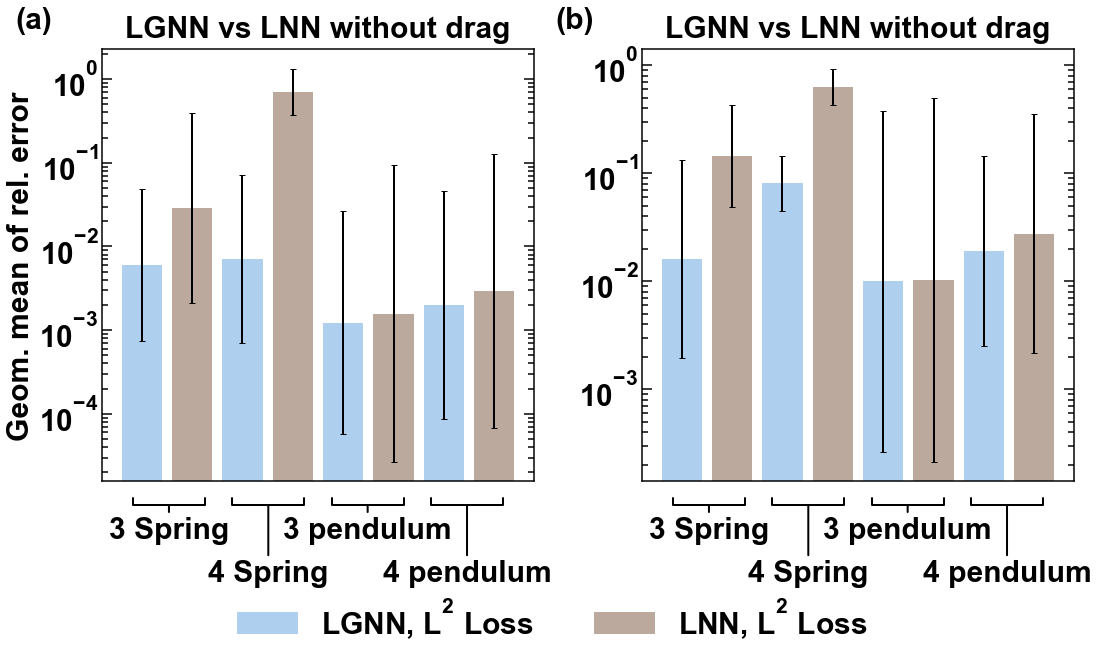}
\end{subfigure}
\begin{subfigure}{\columnwidth}
    \centering
    \includegraphics[width=0.9\columnwidth]{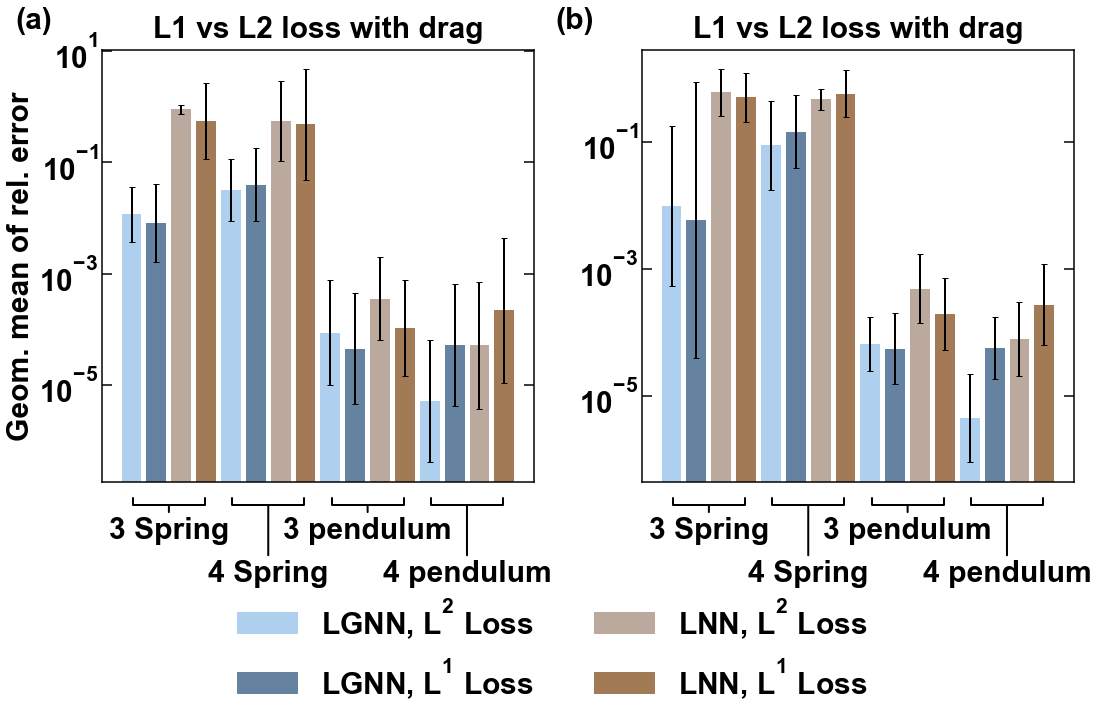}
\end{subfigure}
\begin{subfigure}{\columnwidth}
    \centering
    \includegraphics[width=0.9\columnwidth]{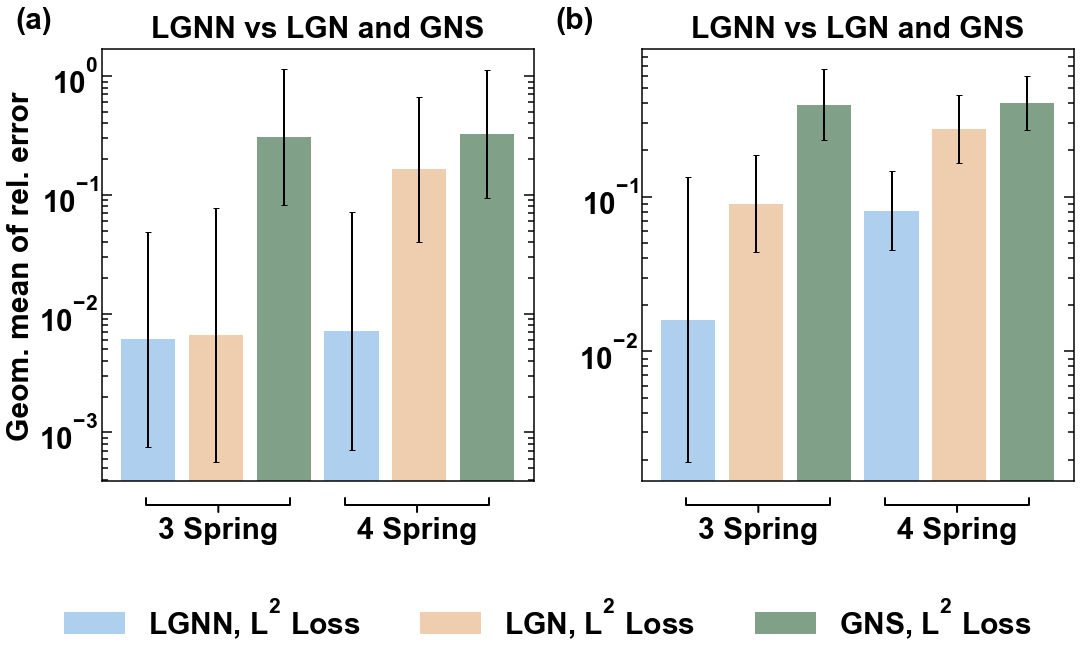}
    \caption{Geometric mean of relative error for (a) rollout and (b) energy for \name and \lnn without drag (top row). We observe that \name consistently exhibits a lower error. (Middle row) $L^1$ vs $L^2$ drag for (a) \name and (b) \lnn. (Bottom row) Comparison of \name with \textsc{Lgn} and \textsc{Gns}. for (a) rollout and (b) energy.}
\end{subfigure}
\end{figure}        





\end{document}